\documentclass{article} 
\usepackage{nips13submit_e,times}
\usepackage{hyperref}
\usepackage{url}
\usepackage[utf8x]{inputenc}
\usepackage[english]{babel}
\usepackage{algorithm}
\usepackage{algorithmic}
\usepackage{graphics}
\usepackage{graphicx}
   \usepackage{tikz}

\def\bbbr{{\rm I\!R}}

\title{Bayesian Optimization for Synthetic Gene Design}
\nipsfinalcopy

\author{Javier Gonz\'{a}lez$^{1,2,3}$, Joseph Longwoth$^{2}$, David C. James$^{2}$ and  Neil D. Lawrence$^{1,3}$ \\
$^1$Department of Computer Science\\
$^2$Department of Chemical and Biological Engineering\\
$^3$ Sheffield Institute for Translational Neuroscience. University of Sheffield \\
\texttt{\{j.h.gonzalez,j.longworth,d.c.james,n.lawrence\}@sheffield.ac.uk} 
}

\begin{document}

\maketitle

\begin{abstract}
We address the problem of synthetic gene design using Bayesian optimization. The main issue when designing a gene is that the design space is defined in terms of long strings of characters of different lengths, which renders the optimization intractable. We propose a three-step approach to deal with this issue. First, we use a Gaussian process model to emulate the behavior of the cell. As inputs of the model, we use a set of  biologically meaningful gene features, which allows us to define optimal \emph{gene designs rules}.  Based on the model outputs we define a multi-task acquisition function to optimize simultaneously severals aspects of interest. Finally, we define an \emph{evaluation function}, which allow us to rank sets of candidate gene sequences  that are coherent with the optimal design strategy. We illustrate the performance of this approach in a real gene design experiment with mammalian cells.
\end{abstract}

\section{Introduction}

Synthetic biology concerns with the design and construction of new biological elements of living systems and the re-design of existing ones for useful purposes \cite{Fremont2012}. In this context, there is a current interest in the development of new methods to engineer living cells in order to produce compounds of interest \cite{export:79443}. A modern approach to this problem is the use of  synthetic genes, which once `inserted' in the cells can modify their natural behavior activating the production of proteins useful for further pharmaceutical purposes.

We present the first approach for gene design based on Bayesian optimization (BO) principles. The BO framework \cite{68paper,journals/corr/abs-1012-2599,osborne2010,HennigSchuler2012} allows us to explore the gene design space in order to provide rules to build genes with interesting properties, such as genes that are able to produce proteins of interest, or genes able to act on the cell lifespan. We use a Gaussian process \cite{Rasmussen:2005:GPM:1162254} to emulate the complex behavior of the cell across the different gene designs. An acquisition function is optimized to deal with exploration-exploitation trade-off. To provide, not only rules for gene design, but current gene sequences candidates, we introduce the concept of \emph{evaluation function}. The goal of this functions is to avoid the bottleneck of optimizing over the sequences by providing rules to rank biologically feasible genes coherent with the obtained gene design rules. 

Although in this work we focus on the optimization of the translational efficiency of the cells, this framework can be generalized to multiple synthetic biology design problems,  such as the optimization of media design, or the optimization of multiple gene knock-out strategies.

\section{Rewriting the Genetic Code}

\begin{figure}[ht]
	\centering
  \includegraphics[width=.7\textwidth]{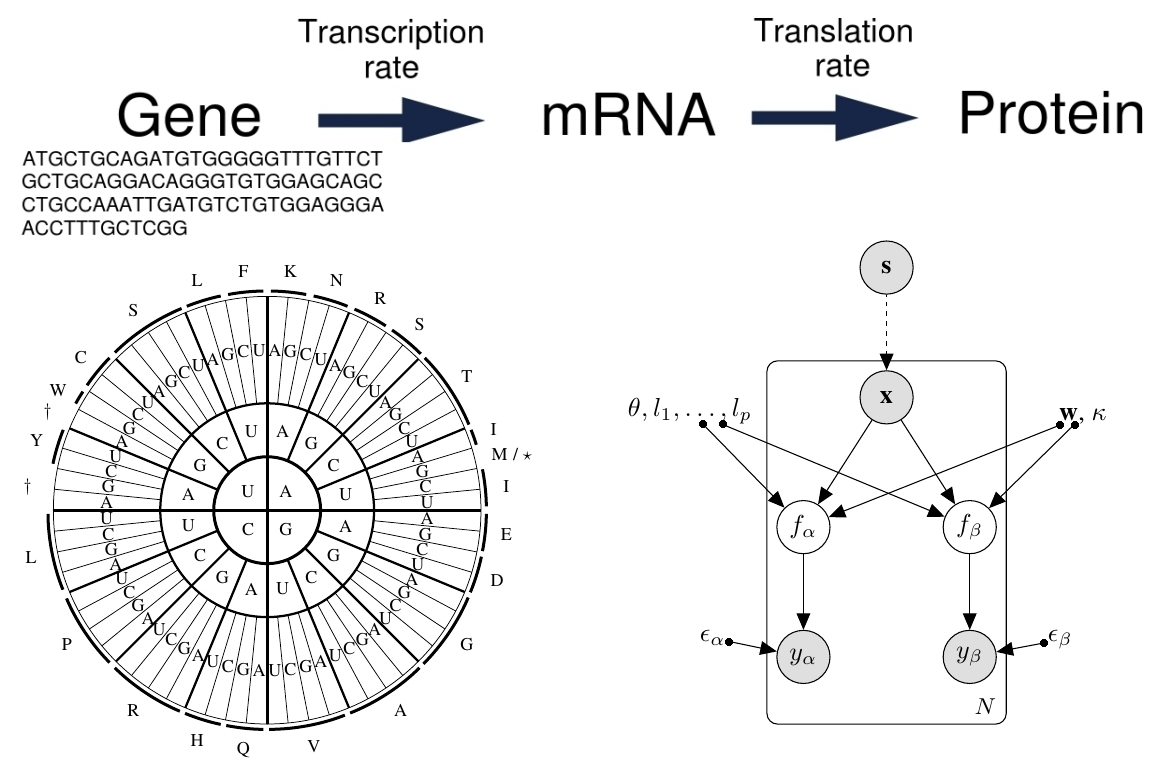}
\caption{\emph{Top}: Central dogma of molecular biology. Each gene contains the information to encode an mRNA molecule which is used by the cell to produce proteins. Both, mRNA molecules and proteins are produced at certain rate. The goal is to design genes sequences able to increase the rates while encoding the same protein. \emph{Bottom left}: Graph of codons redundancy.  Letters inside the circle represent DNA basis and letters outside represent amino acids. The arcs outside the circle cover the paths of redundant codons. $\dagger$ and $\star$ represent special codons.  \emph{Bottom right}: Graphical model of the multi-output Gaussian process used to emulate the cell behavior in this work. }
	\label{fig1}
\end{figure}

Broadly speaking, in molecular biology it is assumed that a gene contains the information to encode a mRNA molecule. The production of such molecules is called \emph{transcription} and takes place in the cell nucleus at certain rate $y_{\alpha}$. Later on, the mRNA molecules are used to produce proteins at a different rate $y_{\beta}$, in what is called the \emph{translation} phase. A one-to-one correspondence between genes and proteins is assumed. Each gene, itself, is also made up of a sequence $\textbf{s}$ of several hundreds of bases $(A,T,G,C)$, triplets of which, form the so-called codons. We can interpret the codons like the `words' in which the genetic code is written. The 64 possible codons encode 20 amino acids, which are the fundamental elements that the cell uses to produce proteins.  This means that the genetic code is redundant: the same aminoacid can be encoded by different codons and therefore there exist multiple ways of encoding the same protein. See Figure \ref{fig1} (top and bottom left) for an illustration of this process.  A fundamental of gene design is that redundant codons choices do not affect the type of protein that is being encoded but they may affect the rates $y_{\alpha}$, $y_{\beta}$, and therefore the efficiency at which it is produced. 

Consider a  p-dimensional representation $\textbf{x} \in \bbbr^{p}$ of a gene sequence $\textbf{s}$. Such a representation will typically be the frequency of the different  codons but it may also include other variables  like the length of the sequence, or the times a certain pattern is repeated across the gene. Denote by $f_{\alpha},f_{\beta}: \mathcal{X} \to \bbbr\times \bbbr$ the functions representing the expected transcription and translation rates given a sequence with features $\textbf{x} \in \mathcal{X}$.  We want to solve the global optimization problem of finding the sequence that maximizes both rates. However, this requires optimization across the all possible sequences. This is infeasible due to the high dimensionality. Instead, we aim to solve the surrogate muti-objective problem of finding  $ \textbf{x}^{\star} = \arg \max_{x \in \mathcal{X}} (f_{\alpha}(\textbf{x}),f_{\beta}(\textbf{x}))$, to later connect $\textbf{x}^{\star}$ with  a particular gene design. 


\section{Method used}\label{seq:model}

\subsection{Multi-output Gaussian Processes as Cell-behavior Surrogate Model}

Let $\{\textbf{s}_1,\dots,\textbf{s}_N\}$ be a set of gene sequences. Consider a p-dimensional feature representation of the sequences given by $\{\textbf{x}_1,\dots,\textbf{x}_N\}$ where $\textbf{x}_i \in \bbbr^{p}$. Let $\textbf{y}_{\alpha},\textbf{y}_{\beta} \in \bbbr^N$ be the observed transcription and translation rates. Our first goal is to learn from a model of the combined data $\mathcal{D} = \{\mathcal{D}_{\alpha},\mathcal{D}_{\beta} \}$, where $\mathcal{D}_{\alpha} = \{\textbf{x}_i, y_{\alpha,i} \}_{i=1}^N$ and $\mathcal{D}_{\beta} = \{\textbf{x}_i, y_{\beta,i} \}_{i=1}^N$, how to predict the value of the output functions  $f_{\alpha}(\textbf{x})$ and $f_{\beta}(\textbf{x})$ at any $\textbf{x} \in \bbbr^{p}$.  For simplicity we assume here that both rates are available for all the sequences, but this assumption can be easily relaxed.

A Gaussian process (GP) is a stochastic process with the property that each linear finite-dimensional restriction is multivariate Gaussian \cite{Rasmussen:2005:GPM:1162254}. GPs are typically used as prior distribution over functions. In the simple output case, the random variables are associated to a single process $f$ evaluated at different $\textbf{x}$ but, this can be easily generalized to multiple outputs, where the random variables are associated to different processes $\{f_l\}_{l=1}^d$. In our case we have $d=2$, which correspond to the two rates of interest. We work therefore with the vector-value function $\textbf{f} :=(f_{\alpha},f_{\beta})$, which is assumed to follow a GP $\textbf{f}\approx \mathcal{GP}(\textbf{m},\textbf{K})$ where $\textbf{m}$ is a 2-dimensional vector whose components are the mean functions $m_{\alpha}$, $m_{\beta}$ of each output and $\textbf{K}$ is a positive matrix valued function that acts directly on input example and tasks indices. The entries  $(\textbf{K}(\textbf{x},\textbf{x}') )_{l,l'}$ in $\textbf{K}(\textbf{x},\textbf{x}')$ represent the covariance between $f_{\alpha}(\textbf{x})$ and $f_{\beta}(\textbf{x}')$. Under a Gaussian likelihood assumption, the predictive distribution for a new vector $\textbf{x}_*$ is taken to be Gaussian such that $p(\textbf{f}(\textbf{x}_*) | \mathcal{D},\textbf{f},\textbf{x}_*,\phi) = \mathcal{N} (\textbf{f}_{*} (\textbf{x}_*),\textbf{K}_{*}(\textbf{x}_*,\textbf{x}_*) )$ where $\textbf{f}_{*} (\textbf{x}_*)$ and $\textbf{K}_{*}(\textbf{x}_*,\textbf{x}_*)$ are close expressions that  depend on the set of input $\textbf{X}$ and the kernel $\textbf{K}$. See \cite{Rasmussen:2005:GPM:1162254,Alvarez:2012:KVF:2344402.2344403} for details. $\phi$ represents all the parameters of the kernel, which can be built following various strategies. In this work we use a combination of the linear and the intrinsic corregionalization models \cite{Teh05,Alvarez:2012:KVF:2344402.2344403}. We take $\textbf{K}(\textbf{X},\textbf{X} ) = \textbf{B}_1 \otimes K_{lin}(\textbf{X},\textbf{X}) + \textbf{B}_2 \otimes K_{se}(\textbf{X},\textbf{X})$, where $K_{lin}$ is a linear kernel used to account for the different levels of the rates and $K_{se}$ a square exponential kernel with a different lengthscale per dimension. $\textbf{B}_{lin}$, $\textbf{B}_{se}$ are the corregionalization matrices, which are parametrized as $\textbf{B}_{lin} =\textbf{w}_{lin}\textbf{w}_{lin}^T + \kappa_{lin}\textbf{I}_2$ and $\textbf{B}_{se} =\textbf{w}_{se}\textbf{w}_{se}^T + \kappa_{se}\textbf{I}_2$ for $\textbf{w}_{lin},\textbf{w}_{se},\kappa_{lin},\kappa_{se}\in \bbbr^{2}$ and $\textbf{I}_2$ is the identity matrix of dimension 2. $\otimes$ represents the Hadamard product. See Figure \ref{fig1} (bottom right) for a graphical description of the model.

\label{alg:bo}
\begin{algorithm}[t!]
\begin{algorithmic}
{ \small
\STATE Extract features $\{\textbf{x}_1,\dots,\textbf{x}_N\}$ from the available gene sequences  $\{\textbf{s}_1,\dots,\textbf{s}_N\}$. 
\STATE Take $\mathcal{D}_{1} = \{\mathcal{D}_{\alpha},\mathcal{D}_{\beta} \}$, where $\mathcal{D}_{\alpha} = \{\textbf{x}_i, y_{\alpha,i} \}_{i=1}^N$ and $\mathcal{D}_{\beta} = \{\textbf{x}_i, y_{\beta,i} \}_{i=1}^N$.
\FOR{$t=1,2,\dots$}
\STATE Fit a multi-output GP model using $\mathcal{D}_{t}$.
\STATE Obtain design rules by taking $\textbf{x}_{t+1} = arg \max_{\textbf{x} \in \mathcal{X}} acqu(\textbf{x}|\mathcal{D}_{t}).$
\STATE Generate a set of candidate gene sequences $\mathcal{S}$.
\STATE  Rank the sequences in $\mathcal{S}$ and select $\textbf{s}_{t+1} = \arg \min_{\textbf{s} \in \mathcal{S}} eval(\textbf{s}|\textbf{x}_{t+1})$.
\STATE Run experiment using $\textbf{s}_{t+1}$ and extract  features $\textbf{x}_{t+1}$ from the sequence $\textbf{s}_{t+1}$. 
\STATE Augment the data $\mathcal{D}_{t+1} = \{\mathcal{D}_{t}, (\textbf{x}_{t+1}, (y_{\alpha,t+1},y_{\beta,t+1})) \}$.
\ENDFOR
\STATE \textbf{Returns}: Optimal gene design $\textbf{s}^{\star}$.
}
\end{algorithmic}\caption{Bayesian optimization for gene design}
\end{algorithm}

\subsection{Acquisition and Evaluation Functions}\label{sec:ev}
In multi-task optimization problems a typical issue is to deal with potential conflicting objectives, or tasks that cannot be optimized simultaneously; in our case this means that both rates cannot be optimized simultaneously using the same sequence. Following previous work in multi-task Bayesian optimization \cite{NIPS2013_5086}, here we focus on an acquisition function that maximizes the average of the tasks. The predictive mean and variance of the average objective are $\bar{m}(\textbf{x}) = \frac{1}{2} \sum_{l=\alpha,\beta}\textbf{f}_*(\textbf{x})$, and $\bar{\sigma}^2(\textbf{x}) = \frac{1}{2^2}\sum_{l=\alpha,\beta}\sum_{l'=\alpha,\beta}(\textbf{K}_*(\textbf{x},\textbf{x}))_{l,l'}. $
Both $\bar{m}(\textbf{x})$ and $\bar{\sigma}^2(x)$ can be used in a standard way using any acquisition function $acq(\textbf{x})$, such as the expected improvement (EI) \cite{68paper}. 

Consider the optimal gene design given by $\textbf{x}^{\star}= arg \max_{\textbf{x} \in \mathcal{X}} acq(\textbf{x})$. Assume that we are interested in the production of a certain protein whose sequence is $\textbf{s}_k$. To improve the sequence $\textbf{s}_k$ according to the optimal design rules $\textbf{x}^{\star}$ without changing the nature of the protein, we can interchange redundant codons, that is, codons encoding the same aminoacid. See Figure 1 (bottom right).  Given a set of sequences satisfying this criteria,  we introduce an evaluation function to rank them in terms of their coherence with the optimal design. In particular we choose  $eval(\textbf{s} | \textbf{x}^{\star})  = \sum_{j=1}^pw_j |\textbf{x}_j-\textbf{x}^{\star}_j|$
where $\textbf{s}$ is a `coherent' sequence, the $\textbf{x}_j$ are the features of $\textbf{s}$, $\textbf{x}^{\star}_j$ are the features of the optimal design and $w_j$ are weights that we choose to be the inverse lengthscales of the $K_{se}$. See Algorithm 1.

\section{Gene Optimization in Mammalian Cells}
\begin{figure}[t!]
	\centering
  \includegraphics[width=.7\textwidth]{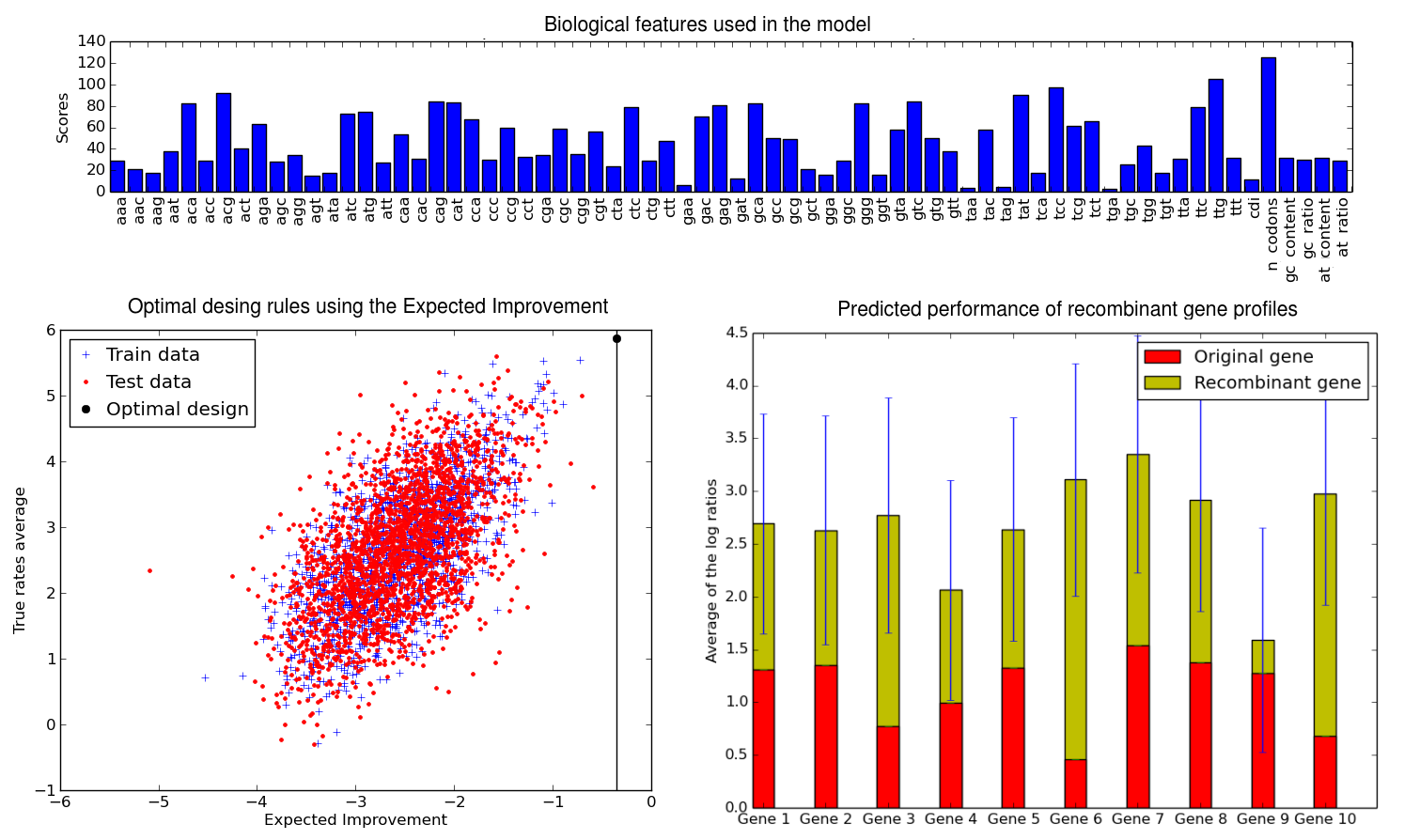}
\caption{\emph{Top}: Inverse lengthscales of the ARD component of the model. \emph{Bottom left}: Optimal design rules selected using EI in the context of the true performance of 3810 genes in mammalian cells.  \emph{Bottom right}: comparison of the true performance of 10 genes with the predicted performance of recombinant genes selected among 1,000 random generated sequences. 95\% confident intervals for the predictions are shown in blue. }
	\label{fig2}
\end{figure}

In this experiment we use Bayesian Optimization to design genes able to optimize the transcription and translation rates of mammalian cells. We use the dataset in  \cite{citeulike:9312122} in which the rates of  3810 genes cells are available. The associated sequences were extracted from http://www.ensembl.org  using the first ENSEBL identifier of the database. As features of the model, we used the frequency of appearance the 64 codons, together with the length of the gene, the GC-content, the AT-content, the GC-ratio and the AT-ratio. We randomly sampled 1500 genes that we used to train the model described in Section \ref{seq:model}. We fit the hyper-parameters of the model by the standard method of maximizing training-set marginal likelihood, using L-BFGS \cite{citeulike:1284223} for 1,000 iterations and selecting the best of ten random restarts. We select the optimal gene design by means of the expected improvement, which we optimize across all the available genes (both train and test sets) in order to have a way of evaluating the coherence of the result with real experimental data. In Figure \ref{fig2} (bottom left) we show the scatter plot of the EI evaluated in all gene features vs. the true average of the ratios. The best possible design is selected by the EI criteria. Next, we select 10 difficult-to-express genes by selecting ten random genes among those whose average log ratio is smaller than 1.5. By taking their sequence as a reference we generated 1,000 random sequences (for each gene) able to encode the same protein. All  across the sequences, we replace each codon with a redundant one, which is sampled uniformly from the set of codons encoding the same aminoacid. Using the evaluation function in Section (\ref{sec:ev}) we ranked the sequences and selected the top rated. in Figure 2 (bottom right) we show the true performance of the sequence (experimental value) versus the predicted value of best recombinant sequences. In the ten cases the recombinant sequence outperforms the original one.   

\section{Conclusions and Challenges}
\vspace{-0.5cm}
We have shown that Bayesian optimization principles can be successfully used in the design of synthetic genes.  One of the most important aspects in this process is to have a good surrogate model for the cell behavior able to lead to appropriate acquisition functions. Considering future models, the fact that the cell is a extremely complex system will be a key aspect to take into account. To optimize certain features of the cell, massive amounts of data will be required, which will require the used of sparse Gaussian processes. Regarding the optimization aspects of the problem, in this work we have worked with a set of features extracted from the gene sequences, which we have used to obtain gene design rules rather than optimal sequences. The use of more features will potentially lead to better and more specific gene design. This will require, however, the development of scalable Bayesian optimization   methods able to work well in high dimensions in the line of some recent works \cite{Wang:2013b,AL,Bergstra+al-NIPS2011,Hutter:2011:SMO:2177360.2177404}.  An alternative approach is to focus directly on the optimization on the sequences rather than on extracted features by omitting any previous biological knowledge. This seems feasible from the modeling point of view by means of the use of string or related kernels \cite{Lodhi:2002:TCU:944790.944799} but the optimization of the acquisition functions derived from this models remains challenging.

\textbf{Acknowledgments}: The authors would like to thank BRIC-BBSRC for the support of this project (No BB/K011197/1).

\small{
\bibliography{bo}
\bibliographystyle{plain}
}

\end{document}